# Rotation Invariance Neural Network

Shiyuan Li

**Abstract** - Rotation invariance and translation invariance have great values in image recognition tasks. In this paper, we bring a new architecture in convolutional neural network (CNN) named cyclic convolutional layer to achieve rotation invariance in 2-D symbol recognition. We can also get the position and orientation of the 2-D symbol by the network to achieve detection purpose for multiple non-overlap target. Last but not least, this architecture can achieve one-shot learning in some cases using those invariance.

## 1 Introduction

Convolutional neural networks have recently shown outstanding performance on image classification and object detection tasks [14]. Have rotation and translation invariance are important goals for model design in image recognition [2, 11, 13]. Although convolutional neural network can achieve some kind of rotation and translation invariance, yet that is based on large train data and can't recognize the samples with rare angles. That is not good enough for real world applications [7] which need almost same performance for same target in different position and orientation, like satellite pictures [3, 4] or microscope pictures [5, 6].

The structure of convolutional neural networks have naturally achieved some level of translation invariance, could this architecture works on rotation invariance? The answer is yes. We designed a new layer in deep convolutional neural networks models called cyclic convolutional layer to solve this problem. The cyclic convolutional layer is mainly a 3-D convolutional layer that treat the feature map as a 3-D array to perform the 3-D convolution, there are two difference between the cyclic convolutional layer and common 3-D convolutional layer. Firstly, the kernel size in the channel dimension is same to feature map size in that dimension. Secondly, the padding in the channel dimension is not zeros but the other side of the feature map (Figure 1a). And the output is also a 4-D array with 2 dimensions for translate and 1 dimension for rotate variance and 1 dimension for different kernels.

This layer can convert rotation variation into a translation variation which is easy to deal with convolution operation, it means that the different orientation will lead to same activation pattern in different position on the 3-D feature map produced by the cyclic convolutional layer. To make this work, every kernel in the previous layer should be rotatable. For example, assume there are $k$ kernels in layer $l$, and the rotation resolution is $r$ degree (so there are $n = 360/r$ different orientation), than every kernel in layer $l$ should have $n$ duplication, each for every orientation. So the different orientation of the symbol will active the same kernel of the different orientation, and the followed cyclic convolutional layer will convert the different orientation into a new dimensions of translation (Figure 1b).

On the other hand, the position and orientation of the object is also important information for real world image process tasks, common convolutional neural networks can only recognize the class of the object and ignore the position and orientation of the object. By add and fully convolutional layer after the cyclic convolutional layer, we can get the position and orientation of the object by a 3-D heat map produce by the last layer of the network.

Last but not the least, human being have the ability to look at an object once and remember it. When you saw someone's picture, you will know the person when you meet him, although the picture only contain one angle and one position of the person. By use the output of the last but one layer of the network as a feature, we can train a linear classify, which use one sample of the object in one position and one orientation as train data and can recognize it in different positions and orientations. By this we achieve some kind of one-shot learning.

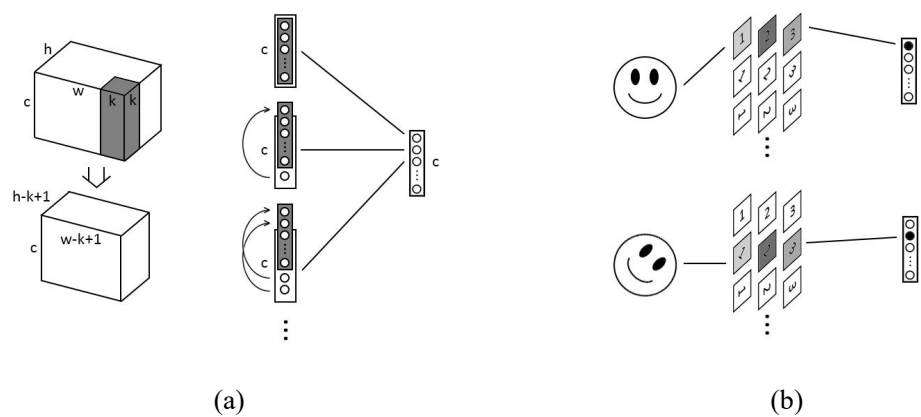

(a)　　　　　　　　　　　　　　　　　　(b)

Figure 1: (a) structure of the cyclic convolutional layer, w, h and c stand for the width, height and channels of the feature maps. (b) Sketch of an image in different angle to active the different filter of a layer.

## 2 Previous Work

Scale-Invariant features are well studied in computer vision and hand craft feature have widely used in many practical applications. SIFT [1] use key points extracted from images and stored in a database, transforms an image into a large collection of feature vectors, each of which is invariant to image translation, scaling, and rotation. Deformable part models [10] with designed features assumes an object is constructed by its parts, the detector first found a match of its whole, and then using its part models to fine-tune the result. Which is limited to a small set of sub-structure, and destroy information that could be used to whole model object.

Neural network-based architectures recently had great success in computer visions and significantly advancing the state of the art on image classification and object detection datasets. Deep Symmetry Networks [8] use symmetry group to describe any form of transformation, turn it into a set and use kernel-based interpolation to compute features through a pooled map to achieve rotation invariance. Scattering networks [12] are cascades of wavelet decompositions designed to be invariant to demonstrate translation and rotation invariance. Rotation-invariant convolutional neural networks [9] rotate the input images in different angles, than compute different images with the same convolutional filters, the output feature maps of those are than concatenated together, and one or more dense layers are stacked on top of it to achieve rotation invariance. Well this don's fully use the different rotated filters.

Our approach for neural network-based rotation invariance is to directly rotate the filter of the convolutional neural network by an affine transformation, and stack the filters in the order of rotated angles, than apply new convolutional layer on top of it, so we can use all of the benefit of rotated filters. All the previous work has some pooling operation through the rotate dimension, this will lost the orientation information of the object, instead of pooling, we use convolutional to maintain the orientation information of the object.

# 3 Model

We first train a very simple Lenet-like [2] convolutional neural network use very little samples (only 15 images for 15 different symbols, to avoid the "6" rotate 180 degree and become a "9" problem, all the symbols are not rotationally symmetric, shown in figure 2) and rotate the filter of convolutional layers to build the rotation invariance neural network. Than test it with a dataset generated by the training set use random rotation and translation.

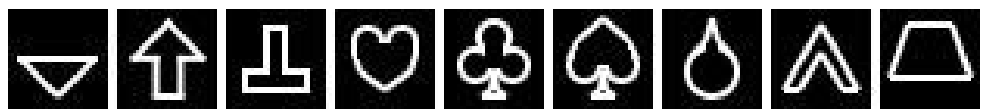

Figure 2: Some of the symbols we use

Our training process is in a greedy way, we first train the network use standard back propagation gradient descent method, than rotate the filter of the first convolutional later and use it to initial a new network. Than we train the new network and fix the first layer's weights, after the network is converged we rotate the filter of the second convolutional layer and use the first and second layer's filters as the initial of another new network and train it. After several layers, we put the cyclic convolutional layer between the convolutional layers and fully connected layers, initial it (with an identity matrix in this paper) and add the fully convolutional layers after the cyclic convolutional layer. Here we consider the fully connected layer as a convolutional layer of kernel size 1x1, and use the same weights of the fully connected layers as the fully convolutional layer's weights. Than we can fine-turning the whole network use back propagation. But in our case it is a very simple network with very little samples, so after the last but one step, we already yield a 100% accuracy in the test dataset so we didn't actually do the last step.

**Rotate the Filter**

The rotate process is a little tricky here, we treat every channel of the filter as a 2-D image and directly rotate the image use an affine transformation. But due to the structure of convolutional neural network, only rotate the filter will lead to the wrong features of the previous layers. The solution we use here has two part (Assuming we have m different filters and will rotate by n different angles, so after the rotate process there will be m*n filters):

1. We arrange the filters (after the rotation operation) by the angles for each different filters (before the rotation operation), it means the first filter (after rotation) is the first filter (before rotation) rotated by 0 degrees (which will produce the first channel of the feature map by the convolution operation of the this filter and the previous layer's feature map), the second filter (after rotation) is the first filter (before rotation) rotated by 360/n degrees (n is the cyclic numbers, stand for how much different angles we process)… and the nth filter (after rotation) is the first filter (before rotation) rotated by 360*(n-1)/n degrees. Than the n+1th filter (after rotation) is the second filter (before rotation) rotated also by 0 degrees. The n+2th filter (after rotation) is the second filter (before rotation) rotated also by 360/n degrees…finally the m*nth filter (after rotation) is the mth filter (before rotation) rotated by 360*(n-1)/n degrees.

2. When we rotate the filter, we also cycle the filter channels, it means we will put the first channel of the filter to the second channel, put the second channel of the filter to the third channel… and put the nth channel of the filter to the first channel. Then put the n+1th channel to the n+2th place, the n+2th channel to the n+3th place…the n+nth channel to the n+1th place.

This two step will solve the wrong features problem. We also fine-turning the network after the rotation use rotated samples.

**Cyclic Convolutional Layer**

When enough layers of filter is rotated, we put the cyclic convolutional layer on top of the it. As mentioned, the cyclic convolutional layer is basically a 3-D convolutional layer that treat the previous layer's feature map as a 3-D feature map of channel size 1, the different between them is the padding in the channel dimension. Just like rotation is cyclic, the cyclic convolutional layer will cycle the rotated channels of the feature map (or the channels of the filter), meanwhile the x and y dimensions are same to the normal 3-D convolution. The cycle is for every rotated group, for example, assume the layer has m different filters and rotated by n different angles (so there are m*n channels of the feature map, and it is arranged by the angles), it will make every rotate group has n channels and there are m groups. The cyclic convolution operation is demonstrate in Figure 1, and this operation will do n times for each angle, every time will produce a feature map of w*h*k, where k is the kernel number of the cyclic convolutional layer. In the end there will be a 4-D array of w*h*k*n, like the normal 3-D convolution.

**Background Depress**

We don't want the background to active any output of our classifier, so we have to depress the background. The usual way to do this is to add another class for background, but this may lead to uneven sample distribution.

Consider that if the input of the network is empty (all zero vector), the reasonable output of the network should also be empty (all zeros before the softmax), but neural network usually don't. How to make this reasonable result happen? The all zero input will make the inner product zero, after subtract the bias, if the result is negative, the output will be zero after the ReLu, so the key to this problem is a negative bias. To achieve this, we add some background samples in the train samples and have all zero labels, the derivation operate of the softmax will direct put all negative gradient to all the former layers, it directly minus the bias. As this is a kind of regularization to the network, add too much of the background sample will make the network not converge. The result heat map is shown in figure 3.

# 4 Experiment

We use 15 32x32 images for training sample to train a network, and random rotate and translate those images to generate 1000 64x64 images for test sets, after the rotate filter step, we achieve 100% accuracy rate, so we accomplish the rotate invariance and it also can consider as a way to achieve one shot learning.

**Implement detail**

The network we use is a Lenet-like network, it has two layer of convolution and two layer of fully connection. The first layer has 6 7x7 filter for 6 orientation for edge detect, it arranged by the angle. After a pooling layer of size 2 there is another convolution layer of 36 14x14 filters, and two fully connected layer of kernel number 36 and 24. We didn't use the second pooling layer because the pooling operation may influence the rotation invariance.

We use 12 for orientation resolution, so there are 36*12 = 432 filters for the second convolution layer, after the cyclic convolution layer, there will be a 3-D feature map for size of 16*16*12, and 36 channels. Than after two fully convolutional layers the final output of the network is a probability map for 15 classes of 16*16*12 pixels. Some of the class's probability map is shown in figure 3.

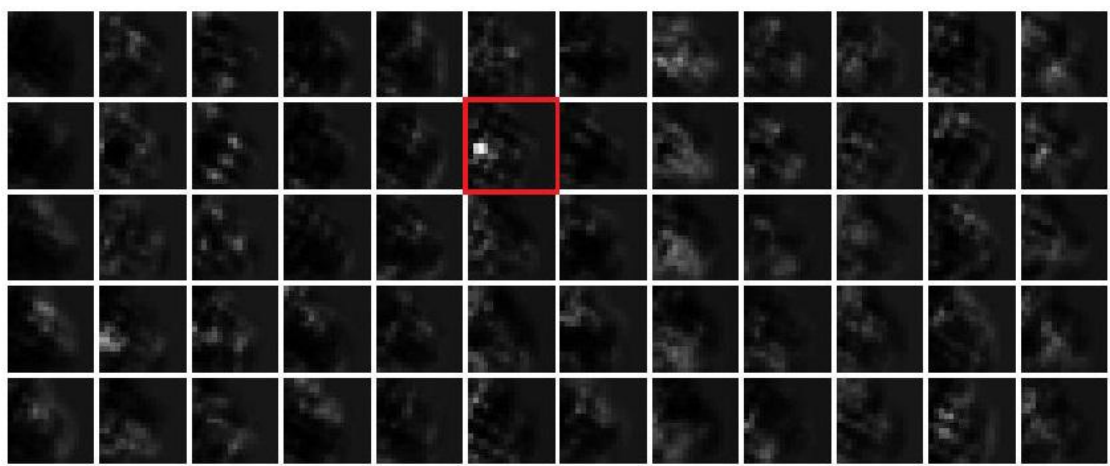

Figure 3: output of the network, each row for one class, each column for one orientation, each sub picture for 16x16 different position. The red square is the final result, for this example the class of the sample is 2, angle is 150º and the symbol is in the middle left of the picture.

**Detection**

Because the output of the network is for each class of each angle and each position, it can output all the different symbols in the same picture, which can use for detection purpose when we put a threshold to distinguish if there is a symbol.

The PR curve of different threshold is shown in figure 4. The number 0.8 is the best threshold in this paper.

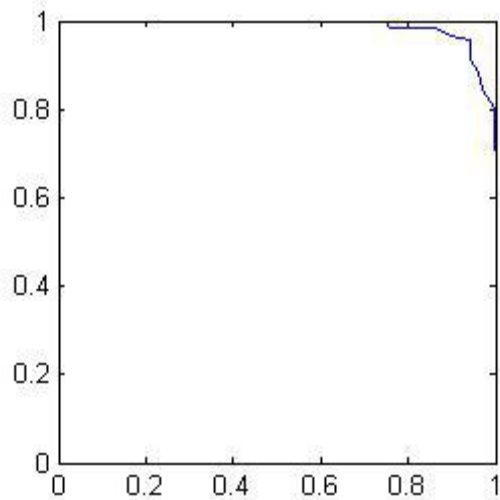

Figure 4: PR curve of detection

**One Shot Learning**

One Shot Learning can use very little sample to train a classifier to recognize new objects [15, 16]. By use the output of the last but one layer as a feature, we can train a linear classify to recognize new samples and add new classes to the network. The positive training data is the new sample of new classes, with the known position and orientation in the 3-D feature map of the last but one layer, and use features near the right position and orientation as negative training data.

## 5 Conclusion

In this paper we bring a method to convert rotation variance into translation variance, and achieve transformation invariance in object recognition. There is other kind of variances like scale and skew, which can also covert into translation variance use the same method in this paper, and it will be a 5-D convolution instead of 3-D convolution. In real world there are 6 free dimensions (3 translate dimensions and 3 rotate dimensions). So it is possible to apply to nature pictures use a 6-D cyclic convolutional layer to achieve transform invariance in real world applications.

The method in this paper will make the network wider instead of deeper, which is not same to the current results in other papers, but shallow network will be fast to compute when the parallel computing resource is abundant, which is almost infinite when computers will be way more cheaply in the future.